\documentclass[conference]{IEEEtran}
\usepackage{cite}
\usepackage{amsmath,amssymb,amsfonts}
\usepackage{algorithmic}
\usepackage{graphicx}
\usepackage{textcomp}
\usepackage{xcolor}
\usepackage{hyperref}
\usepackage{booktabs}
\usepackage{comment}
\usepackage{enumitem}
\usepackage{balance}
\def\BibTeX{{\rm B\kern-.05em{\sc i\kern-.025em b}\kern-.08em
    T\kern-.1667em\lower.7ex\hbox{E}\kern-.125em}}

\begin{document}

\title{Cross-Modal Entity Matching for\\ Visually Rich Documents}

\author{\IEEEauthorblockN{Ritesh Sarkhel}
\IEEEauthorblockA{\textit{Amazon}\\
Seattle, USA}
\and
\IEEEauthorblockN{Arnab Nandi}
\IEEEauthorblockA{\textit{The Ohio State University}\\
Columbus, USA}
}

\thispagestyle{plain}
\pagestyle{plain}

\maketitle

\begin{abstract}
Visually rich documents (e.g. leaflets, banners, magazine articles) are physical or digital documents that utilize visual cues to augment their semantics. Information contained in these documents are ad-hoc and often incomplete. Existing works that enable structured querying on these documents do not take this into account. This makes it difficult to contextualize the information retrieved from querying these documents and gather actionable insights from them. We propose \texttt{Juno} -- a cross-modal entity matching framework to address this limitation. It augments heterogeneous documents with supplementary information by matching a text span in the document with semantically similar tuples from an external database. Our main contribution in this is a deep neural network with attention that goes beyond traditional keyword-based matching and finds matching tuples by aligning text spans and relational tuples on a multimodal encoding space without any prior knowledge about the document type or the underlying schema. Exhaustive experiments on multiple real-world datasets show that \texttt{Juno} {generalizes} to heterogeneous documents with diverse layouts and formats. It {outperforms state-of-the-art baselines} by more than 6 F1 points with up to 60\% less human-labeled samples. Our experiments further show that \texttt{Juno} is a computationally robust framework. We can train it only once, and then adapt it dynamically for multiple resource-constrained environments without sacrificing its downstream performance. This makes it suitable for on-device deployment in various edge-devices. To the best of our knowledge, ours is the first work that investigates the information incompleteness of visually rich documents and proposes a \textit{generalizable, performant and computationally robust} framework to address it in an end-to-end way.\looseness=-1
\end{abstract}


\section{Introduction}
A visually rich document (\texttt{VRD}) is a physical or digital document that leverages explicit or implicit visual cues~(e.g. color, distance, orientation) to augment its semantics. From medical intake forms to invoices, restaurant menus to leaflets, \texttt{VRD}s are pervasive in our everyday lives. Due to their popularity, there is a recent surge in research interest on structured querying~\cite{majumder2020representation,sarkhel2019visual,sarkhel2021improving,xu2020layoutlmv2} of these documents. A data pipeline set up for this 
task typically works as follows. Given a schema~$\mathcal{S}$ and a document~$\mathcal{D}$, we extract a structured record~$\mathcal{R^S_D}$ with schema~$\mathcal{S}$ from~$\mathcal{D}$. We clean this record, transform it into a compatible format and load it onto a data-warehouse. This data-warehouse serves as the back-end of an analytical engine that allows us to execute queries and gather actionable insights from~$\mathcal{D}$. Unfortunately, most existing methods make a \textit{closed-world assumption} in setting up this pipeline that leads to limited query coverage. Take the following scenario for example.\vspace{0.02cm} 
 
\noindent \textbf{Example: }\textit{Alice} wants to place an order from a restaurant she is visiting for the first time. She has a printed menu which contains various items along with their prices. \textit{Alice} has some food allergies. The allergen information of an item, however, does not appear on the menu. This leads to her pouring over the menu and looking up each item in a nutritional table before she can place an order. Existing data pipelines that enable structured querying of visually rich documents cannot help her automate this process as they only support those queries that return a subset of text spans appearing on the document. Due to the ad-hoc and often incomplete nature of these documents, this can be limiting in many real-world scenarios. The information needed to gather insights (e.g. allergen information) may not appear in the document in the first place.\looseness=-1\vspace{0.02cm} 

\begin{figure}[t]
    \centering
    \includegraphics[width=\linewidth]{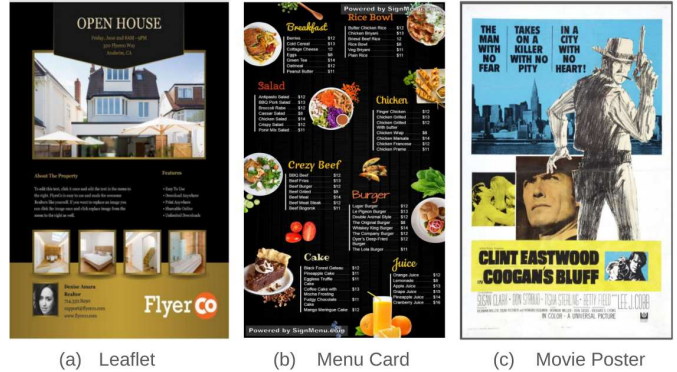}
    \caption{Visually Rich Documents utilize visual and textual cues to highlight the semantics of various entities appearing on them. They can have diverse layouts, formats, and be used for short-form communication such as leaflets, posters and menu-cards.\looseness=-1}\label{fig:1_1}
\end{figure}

\begin{figure*}[t]
    \centering
    \includegraphics[width=0.95\linewidth,height=250pt]{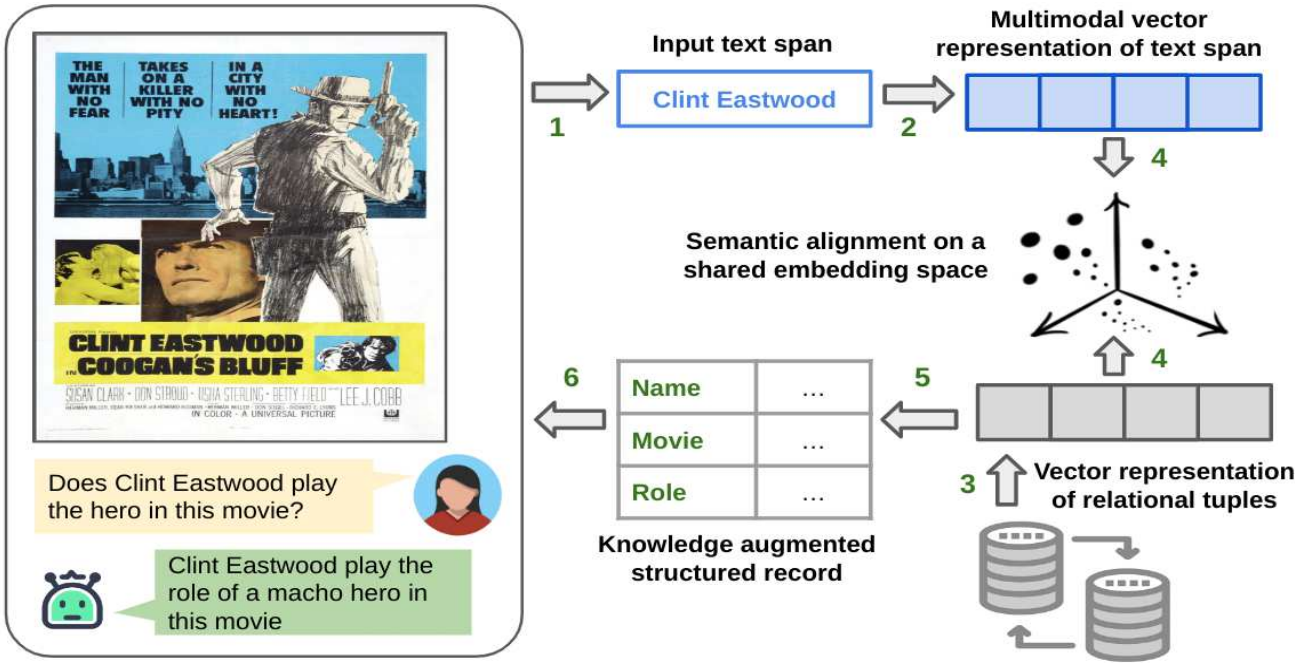}
    \caption{An overview of \texttt{Juno}'s end-to-end workflow is shown on the right side of this figure. It takes a text span from a visually rich document as input, encodes it as a fixed-length vector on a multimodal embedding space, and aligns it with semantically similar tuples in a relational database on that space. These tuples are then retrieved and returned back to the user.\looseness=-1}\label{fig:1}
\end{figure*}

\noindent \textbf{Cross-modal entity matching: }A data pipeline augmented with {on-device cross-modal entity matching} capability (defined in Section~II) can help \textit{Alice} automate this task. Briefly, a cross-modal entity matching~(\texttt{CEM}) framework maps a data element~${e}_1$ in one modality (e.g. text span on a printed menu) to a data element~${e}_2$ in another modality (e.g. tuple in a nutritional database) if they represent the same real-world object. A typical entity matching framework works in two phases. In the first phase, data elements that are unlikely to match are discarded~\cite{papadakis2020blocking}. Remaining data elements from both sources are then matched by making pairwise comparisons through carefully designed rules in the second phase. What makes this task challenging in a cross-modal setting is matching data elements \textit{across modalities in a generalizable way}. Contrary to its unimodal counterpart~\cite{christophides2020overview}, cross-modal entity matching frameworks require an additional {relationship modeling} step~\cite{wang2016comprehensive} before the matching phase, where data elements from different modalities are represented on a shared embedding space. This is a challenging task for \texttt{VRDs} as text spans appearing on these documents employ both visual and textual cues to highlight their semantics. Diversity in the layout and format of these documents also makes it difficult to implement a generalizable solution for this task (see Fig.~\ref{fig:1_1}).\looseness=-1\vspace{0.02cm} 

\noindent \textbf{Limitations of existing solutions: }Before describing our framework, let's walk through a naive solution for the previous example first. Using off-the-shelf data pipelines, we can extract structured records corresponding to each item in the printed menu by employing a document understanding model~(e.g. LayoutLMv2~\cite{xu2020layoutlmv2}). Briefly, it is a Transformer-based~\cite{vaswani2017attention} model that takes a rendered image of the document as input, encodes both its visual and textual features, and extracts structured records corresponding to each item appearing on this document. We can identify the allergen information of an item by performing a fuzzy-join between its corresponding record and a nutritional database. Unfortunately, this solution does not scale well for a large-scale corpus of heterogeneous documents. Performance of a document understanding model is directly proportional to the number of human-labeled samples used to fine-tune~\cite{pan2009survey} that model. Performing fuzzy-join between a record and a relational database requires significant {domain-expertise} as well. For example, an item printed as \textit{`Pasta in Red Tomato Sauce'} on the menu may appear as \textit{`Pasta in Marinara Sauce'} in the database. Establishing semantic similarity in such scenarios requires {carefully designed rules} from domain-experts. It is hard to maintain and update these rules for a large-scale corpus. Off-the-shelf tools, including large language models trained on huge amount of textual data (e.g. GPT-3~\cite{floridi2020gpt}) cannot bridge this gap completely as they hallucinate on emergent topics~\cite{azamfirei2023large,ye2023cognitive} typically covered by these documents (see Section~V for experimental results). Fine-tuning these models on a custom domain requires significant effort and computational resources. Vision-language models (e.g. CLIP~\cite{radford2021learning}) trained on huge amounts of \texttt{\{image, text\}} pairs exhibit good zero-shot generalization capability for cross-modal entity matching and retrieval tasks. These models, however, use pixel-level encodings to represent an image. This does not take the inherently multimodal nature of a visually rich document into account (see Section~V for experimental results). Recent works~\cite{sarkhel2019deterministic,majumder2020representation} have established the necessity of encoding both textual and visual features pertaining to the document layout to compute fixed-length representations for visually rich documents. Furthermore, the amount of computational resources needed to infer a match using these models makes it difficult to deploy them in resource-constrained environments. This motivates us to formalize the following objectives for our entity matching task.\looseness=-1\vspace{0.02cm} 

\noindent \textbf{Problem statement: }Given a document~$\mathcal{D}$ and a relational table~$\mathcal{T}$, our goal is to learn a mapping~$f:w\rightarrow t$ between a text span~$w\in \mathcal{D}$ and a set of tuples~$t\in \mathcal{T}$ if they represent the same real-world object. Our objectives are as follows.\looseness=-1

\begin{enumerate}
    \item $f$ is \textit{performant} i.e., its mapping accuracy is high.\vspace{0.05cm}

    \item $f$ is \textit{scalable} i.e., the number of human-labeled samples needed to learn $f$ is low.\vspace{0.05cm}

    \item $f$ is \textit{generalizable} i.e., it can be adapted for diverse documents without any prior knowledge about the document type or the underlying schema.\looseness=-1\vspace{0.05cm}

    \item $f$ is \textit{computationally robust} i.e., it can adapted to resource-constrained environments without any significant degradation in its downstream performance.
\end{enumerate}

\noindent \textbf{Our contributions: }We develop a generalizable framework for cross-modal entity matching against visually rich documents in this paper. Our core contribution is a multimodal deep neural network that maps text spans with semantically similar relational tuples by aligning them on a shared embedding space without any prior knowledge about the document type or the underlying schema. 
Contrary to existing works that use handcrafted rules from domain experts, our framework is more {scalable} as it leverages a novel attention mechanism~\cite{chorowski2015attention} (defined in Section~II) to reduce the number of pairwise comparisons to find a match. Compared to contemporary supervised solutions, our framework uses significantly less number of human-labeled samples in its training. 
This frees up developers from tedious feature engineering and allows them to focus more on training iterations. Not being bound to layout and/or format-specific rules makes our method \textit{more generalizable} for diverse document types as well. We refer to this framework in the rest of these documents as \texttt{Juno}.\vspace{0.02cm} 

\noindent \textbf{Summary of results: }We evaluate \texttt{Juno} on {two real-world datasets} -- The IMDB Movie Dataset and The NYC Open Event Dataset for separate entity matching tasks. Our results show that \texttt{Juno} is not only \textit{more performant} than existing methods -- outperforming state-of-the-art baselines by more than 6\% in F1-score but it is also \textit{more scalable} -- reducing the number of human-labeled samples needed to train the same model for comparable performance using direct supervision by up to 60\%. Our experiments further show that \texttt{Juno} is \textit{computationally robust} -- we can reduce its memory footprint by up to 24\% using off-the-shelf algorithms without degrading its performance. This makes \texttt{Juno} a suitable candidate for on-device deployment of various cross-modal entity matching tasks. Investigating optimal ways to reduce 
memory footprint and computational overhead of entity matching frameworks for interactive applications (e.g. augmented/mixed reality settings~\cite{burley2022quill}) is one of our planned future works.\looseness=-1
\section{Background \& Related Works}

\subsection{Entity Matching}

\noindent Let, $D_1$ and $D_2$ represent two unique data sources. Entity matching refers to the task of identifying all data element pairs $<e_{1},e_{2}>$, $\forall e_{1}\in D_1$, $e_{2}\in D_2$ that represent the same real-world objects. A typical entity matching framework works in two phases. In the \textit{blocking phase}, it filters out obvious non-matches from both data sources to reduce the number of pairwise comparisons, whereas in the \textit{matching phase} it identifies similar data elements from the remaining set. If $D_1$ and $D_2$ are structured data sources, this can be done by making attribute-wise comparisons between tuples from $D_1$ and $D_2$. If they have the same schema, this is a trivial task. Otherwise, we perform schema-matching~\cite{rahm2001survey,fagin2009clio} to align these two data sources first. Contemporary researchers have investigated various techniques~\cite{smith2022lillie,cafarella2009data} to perform entity matching between a document and a structured data source in recent years. Smith et al.~\cite{smith2022lillie} employed carefully designed rules to identify \texttt{\{subject, predicate, object\}} triplets from the document first. To perform entity matching, they aligned the predicates extracted from these triplets and tuples in a relational database using high-precision rules. Cafarella et al.~\cite{cafarella2009data} employed human-experts to modify, extend, and align web-tables extracted from semi-structured web-pages for data integration tasks. Unfortunately, rule-based solutions like these are hard to scale for a large-scale corpus due to the diversity in layout and format of visually rich documents. Contemporary researchers have proposed deep learning based solutions to improve the scalability of this task in recent years.\looseness=-1

\subsection{Deep Entity Matching}

\noindent A deep entity matching framework usually works as follows. After the {blocking phase} has pruned obvious non-matches, it formulates the {matching phase} as a binary classification problem, where it represents each data element (from both sources) as a fixed-length vector and categorizes each candidate pair as match (or non-match). Employing deep neural networks for this task obviates the need of careful feature engineering to represent a data element, and frees up developers to focus on training iterations. The choice of architecture usually depends on the dataset and the available computation budget. For instance, Ebraheem et al.~\cite{ebraheem2018distributed} developed a Recurrent Neural Network~(RNN)-based~\cite{sutskever2013training} model to match two relational databases by learning a distributed representation for each tuple. Nie et al.~\cite{nie_han_he_sun_chen_zhang_wu_kong_2019} extended their work to develop a label-efficient model that leverages the power of transfer learning~\cite{wu1992fast}. Recent works~\cite{li_li_suhara_doan_tan_2020,zhao2019auto} have established the efficacy of Transformer-based models~\cite{vaswani2017attention} pretrained on large amounts of textual data for this task, reporting state-of-the-art results on multiple benchmark datasets.\looseness=-1

\subsection{Deep Entity Matching with Attention}
\noindent Deep neural networks such as RNN~\cite{sutskever2013training} assigns equal weight to the entire input sequence when computing a fixed-length representation of a data element. This makes it difficult for an entity matching model to learn a meaningful summarization of long and potentially noisy input sequences (e.g. multi-valued attributes in relational tuples). Recent works~\cite{sutskever2014sequence,cho2014learning} have introduced attention mechanism to overcome this limitation. A deep neural network with attention takes a fixed-length vector representation~($y$) of a data element as input, and produces a summarized version~($z$) of~$y$ as output. In doing so, it only retains the information that is relevant to the downstream task and discards the rest thus allowing an entity matching model to `attend' to the `important parts' of an input sequence during the matching phase. Importantly, attention mechanism allows an entity matching model to learn which parts of an input sequence to attend during training. Although not for visually rich documents, Mudgal et al.~\cite{mudgal2018deep} was one of the early works to establish the efficacy of attention mechanism for entity matching tasks. Li et al.~\cite{li_li_suhara_doan_tan_2020} extended these findings and established the efficacy of attention mechanism for entity matching tasks with Transformer-based models in recent years. One of the key differences of the attention mechanism employed in \texttt{Juno} compared to existing works is its asymmetric nature. Contrary to existing works that utilize a symmetric attention mechanism due to the homogeneity of both data sources, we employ an asymmetric, bi-directional attention mechanism that adapts to the diverse characteristics of a candidate pair: multimodal text spans from heterogeneous documents, and tuples from relational databases.

\subsection{Cross-Modal Entity Matching}

\noindent Contrary to its unimodal counterpart, a cross-modal entity matching (\texttt{CEM}) framework matches data elements across modalities. A traditional \texttt{CEM} framework works in three phases~\cite{wang2016comprehensive}. Fixed-length vectors are computed to represent data elements from both data sources in the first phase. These vectors are then projected on a shared, multimodal embedding space in the second phase. Finally, correlation between similar data elements are established on this shared embedding space in the third phase. Researchers have investigated various techniques to learn a shared embedding space that aligns matching data elements from different modalities. For example, Rasiwasia et al.~\cite{rasiwasia2010new} and Sharma et al.~\cite{sharma2012generalized} leveraged canonical correlation analysis and bilinear modeling to learn a {common subspace} that maximizes the correlation between similar data elements. Wu et al.~\cite{wu2010learning} and Carvalho et al.~\cite{carvalho2018cross} followed a metric learning approach to learn a common representation and a similarity threshold across modalities. Cao et al.~\cite{cao2016correlation} and Lin et al.~\cite{lin2015deep} employed hashing-based techniques to learn common representations on a Hamming space. Radford et al.~\cite{radford2021learning} was one of the early works to train a Transformer-based vision-language model on semantically similar \texttt{\{image, text\}} pairs collected from the internet. Alayrac et al.~\cite{alayrac2022flamingo} extended their work to show state-of-the-art results for vision-langauge tasks in zero-shot and few-shot settings. Girdhar et al.~\cite{girdhar2023imagebind} established the efficacy of learning a shared embedding space that encodes multiple modalities simultaneously for cross-modal retrieval tasks.\looseness=-1\vspace{0.05cm} 

\begin{figure*}[t]
    \centering
    \includegraphics[width=0.95\linewidth]{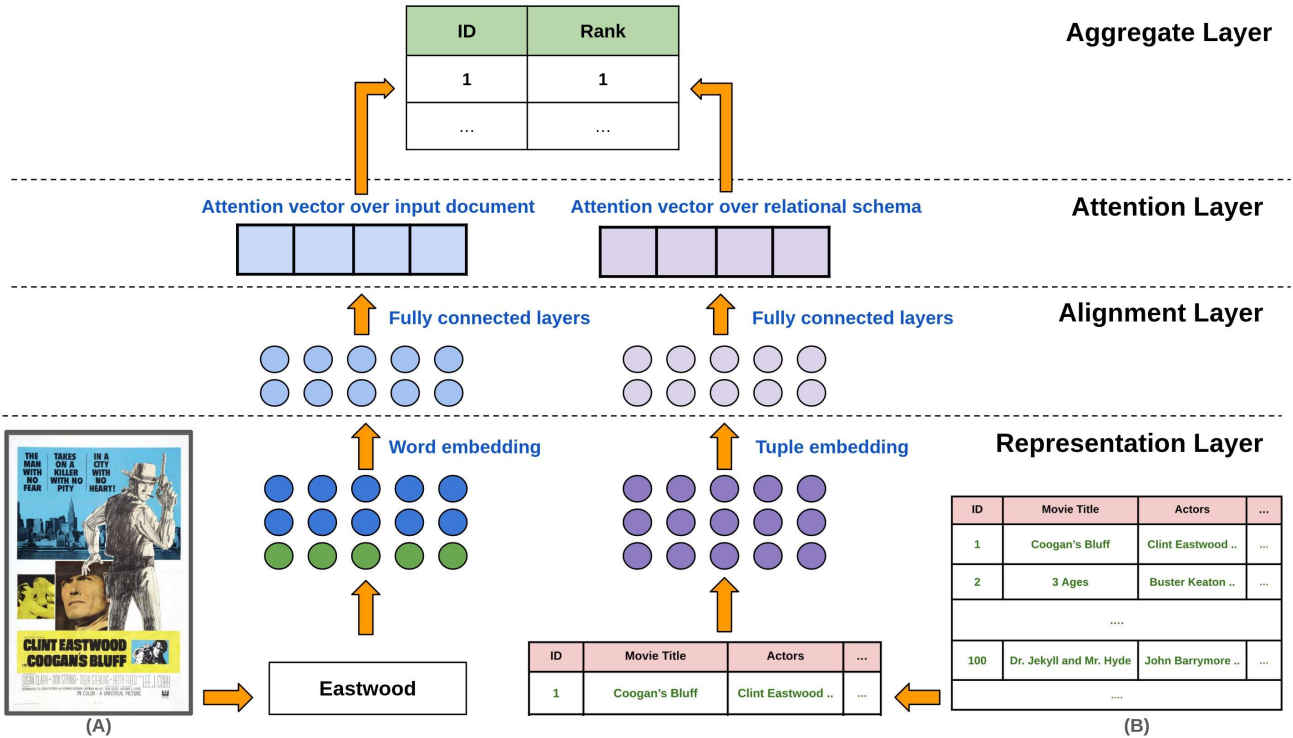}
    \caption{An overview of our neural network architecture. In this example, the network maps a visually rich movie poster (A) to tuples in a relational table (B) containing supplementary information about the movie.\looseness=-1}\label{fig:1}
\end{figure*}

\noindent These methods, however, do not address some of the key challenges of our task. \textit{First}, existing works do not address one of the core issues of aligning a visually rich document against an enterprise-scale database which is scalability. Most existing methods utilize handcrafted rules designed using prior knowledge about the document type, or the underlying schema for this task. This is hard to scale for a large-scale corpus of heterogeneous documents. \textit{Second}, it has been established~\cite{azamfirei2023large,ye2023cognitive}that large-scale models that have been trained on huge amounts of textual and/or multimodal data tend to hallucinate on emergent topics that are not sufficiently represented in their training corpus. Fine-tuning these models on a custom domain require significant effort and computational resources. \textit{Third}, existing works that learn a shared embedding space for cross-modal entity matching/retrieval tasks leverage pixel-level image encodings. Contemporary researchers~\cite{sarkhel2019deterministic,majumder2020representation} have showed the superiority of leveraging layout-based hierarchy to learn a fixed-length representation for visually rich documents in recent years. We address this challenge by developing a data model that represents heterogeneous \texttt{VRDs} and relational tuples in a principled way. We discuss it next.\looseness=-1  
\section{Our Data Model}
\noindent We represent a relational database with schema~$\mathcal{S}$ as a set of tuples~$\{t_1,t_2...t_T\}$. Each tuple~$t_i$ is represented as a nested set~$\{a_{i,1},a_{i,2}...a_{i,n}\}$ where~$a_{i,j}$ is a sequence of tokens denoting the value of the $j$\textsuperscript{th} attribute of schema~$\mathcal{S}$ in tuple~$t_i$.\looseness=-1\vspace{0.05cm} 

\noindent Similarly, we represent a \texttt{VRD} as a nested set~$\{V,H\}$, where $V$ denotes the set of \textit{atomic elements} and~$H$ denotes their \textit{visual organization}. We define them as follows.\looseness=-1

\subsection{Atomic Elements}
\noindent An atomic element represents the smallest visual element in a document. Each {visual span} in a document comprises of one or more atomic elements. We classify each atomic element into two major categories: \textit{text elements} and \textit{image elements}.\vspace{0.05cm}

\noindent \textbf{A.1. Text element:} A text element is a visual element with a text-attribute. We deem each word in a document as a text element in our data model. We represent each text element~$a_{text}$ as a nested set: $a_{text} = \{text$-$data,x,y,w,h\}$, where $text$-$data$ represents the transcription of the span covered by this text element. $h$ \& $w$ denote the height and width of the smallest bounding-box enclosing this text span, and $x,y$ represent the coordinates of its top-left corner. We identify text elements in a document using Tesseract~\cite{smith2007overview}, a popular open-source OCR engine.\vspace{0.05cm}
    
\noindent \textbf{A.2. Image element:} An image element, on the other hand, denotes an image-attribute in the document. We represent an image element~$e_{img}$ as a nested set: $e_{img} = \{pixel$-$data,x,y,w,h\}$, where $pixel$-$data$ represents the pixel-values in the smallest bounding box enclosing~$e_{img}$.\looseness=-1  

\subsection{Visual Organization}
\noindent We represent the visual organization of a document using a tree-like structure~$H$. Each node in~$H$ represents a text span. We define~$H$ using five levels of layout hierarchy following the hOCR specification format~\cite{breuel2007hocr}. In this format, a document is deemed to be made up of several {columns}, a {column} is made up of several {paragraphs}, a {paragraph} is composed of several {text-lines}, and a {text-line} consists of multiple {words}. A node in~$H$ is a child of another node if its text span is enclosed by the text span represented by its parent node. We use an open-source page segmentation algorithm~\cite{smith2007overview} to construct the layout-tree~$H$ in our experiments.\looseness=-1
\section{Methodology}
\noindent At the core of our framework is a multimodal neural network that maps a text span from a visually rich document with semantically similar tuples from a relational database without any prior knowledge about the document type or the underlying schema. It works in two phases.\looseness=-1

\subsection{Phase I: Encoding inputs into fixed-length vectors}

\noindent \textbf{A.1. Encoding text spans:} The first layer of our network, called the \textit{representation layer} computes a distributed representation of a text span in the document. It represents each word~$w$ represented as a leaf node in the document layout-tree as a vector~$\mathcal{F}_w$ of dimensions~$4\times 768$. The first three rows in~$\mathcal{F}_w$ represents the embedding vector encoding~$w$, its immediate bi-gram, and tri-gram respectively. We use the publicly available LayoutLMv2~\cite{xu2020layoutlmv2} model to compute these vectors. More specifically, we average the output from the last two layers of a LayoutLMv2\textsubscript{BASE} model pretrained on the IIT-CDIP dataset~\cite{harley2015evaluation} to compute these vectors. The last row of~$\mathcal{F}_w$ represents an embedding vector from the last fully-connected layer of a pretrained MobileNet\footnote{the MobileNet architecture has been shown~\cite{sarkhel2019deterministic} to be effective in encoding discriminative properties of \texttt{VRDs}\looseness=-1} model~\cite{howard2017mobilenets} from a rendered image of the document. While the first three rows in~$\mathcal{F}_w$ encodes the local context information of the word~$w$, the last row encodes document-level visual cues such as layout, orientation, and formatting of the document.\looseness=-1\vspace{0.07cm}

\noindent \textbf{A.2. Encoding relational tuples:} The representation layer also computes a two-dimensional vector~$\mathcal{F}_t$~($n\times 768$) to represent a relational tuple~$t$. The $i$\textsuperscript{th} row in~$\mathcal{F}_t$ denotes the embedding vector of the~$i$\textsuperscript{th} attribute in tuple~$t$. Following prior works~\cite{nie_han_he_sun_chen_zhang_wu_kong_2019}, we represent each attribute in~$t$ as a sequence of tokens and utilize a pretrained RoBERTa\textsubscript{BASE} model to compute an embedding vector for each attribute. We impute missing values in a tuple with a special token \texttt{[UNK]} {from the model's vocabulary}. For multi-valued attributes, we linearize the attribute values as a long sequence first, and then compute an embedding vector of that sequence. The main reason behind using pretrained models in the representation layer of our network is to leverage the power of transfer learning~\cite{pan2009survey} and minimize the amount of human-labeled samples needed to learn a fixed-length encoding of a data element. Transformer-based models~\cite{vaswani2017attention} such as RoBERTa and LayoutLMv2 are good candidates for this task as they are capable of embedding out-of-vocabulary words with a fixed vocabulary size due to their subword encoding capabilities~\cite{sennrich2016neural}. Other models with such capabilities can also be used instead as they have a transitive effect on the subsequent layers of our network. We establish this flexibility provided by the plug-and-play nature of our architecture through experiments in Section~V.\looseness=-1

\subsection{Phase II: Aligning text spans and relational tuples}

\noindent The second layer of our network, called the \textit{alignment layer} computes pairwise similarities between text spans in the document and relational tuples in the database. It comprises of two fully-connected layers, each with a dimension of~$768\times 768$. The alignment layer takes the vectors~$\mathcal{F}_w$ and $\mathcal{F}_t$ computed by the representation layer as input, and projects them on to a shared embedding space. Let, $\mathcal{F'}_w$ \& $\mathcal{F'}_t$ represent these projected vectors. We compute the distance between a text span~$w$ in the document and a relational tuple~$t$ on this shared space as follows.

\begin{equation}
    \mathbf{L}_{align}(w,t) = argmin_{\hspace{0.05cm}i,j} |\hspace{0.05cm} \mathcal{F'}_w[j] - \mathcal{F'}_t[i] \hspace{0.05cm}|
\end{equation}\label{eq:1}

\noindent To identify a matching tuple~$t^*$ for the text span~$w$, we minimize the distance~$\mathbf{L}_{align}(w,t)$ between $w$ and all tuples~$t$ in the relational database.\looseness=-1

\begin{equation}
    t^* = argmin_{\hspace{0.05cm}t,i,j} |\hspace{0.05cm} \mathcal{F'}_w[j] - \mathcal{F'}_{t}[i]\hspace{0.05cm}|,\forall t
\end{equation}\label{eq:2}

\noindent Unfortunately, this results in a linear scan over the entire database for each text span in the document. This is a major computational bottleneck in terms of scaling our framework up for enterprise-scale databases and verbose documents. This can be mitigated by pruning off unlikely matches from both data sources before any pairwise comparison takes place. Contrary to existing methods that require carefully designed rules from domain-experts for this task, we perform this pruning operation in an end-to-end trainable fashion, with significantly less human-effort. The key enabler in this is the \textit{attention layer} in our network.\vspace{0.1cm}\looseness=-1

\noindent \textbf{B.1. Bi-directional attention for faster alignment: }
Let's assume that the $i$\textsuperscript{th} attribute of tuple~$t$ has the minimum distance from a text span~$w$ in Eq.~2. The attention layer indexes this information for each document in the training corpus using two vector-stores~${V_D}$ and ${V_T}$. ${V_D}$ indexes the embedding vector representing the text span~$w$ (i.e. $\mathcal{F}_w$) and~${V_T}$ indexes the embedding vector for the $i$\textsuperscript{th} attribute in~$t$ (i.e. $\mathcal{F}_t[i]$). Both vector-stores use the~$i$\textsuperscript{th} attribute in~$t$ as the indexing attribute. Once the indexes have been constructed, we cluster the embedding vectors against each indexing attribute using the DBSCAN algorithm. We update both vector stores using the cluster centroids. We only keep the cluster centroids for each indexing attribute. This helps us impose an upper bound on the computational cost of this pruning step, making our inference cost more manageable. Once the vector stores have been updated, we are ready to prune off unlikely matches from our search space. We achieve this by computing two vectors -- one over the input document using the vector store~${V_D}$, and another one over the relational database using the vector store~${V_D}$. We describe how these vectors are computed next.\looseness=-1\vspace{0.1cm}

\begin{figure}[t]
    \centering
    \includegraphics[width=\linewidth]{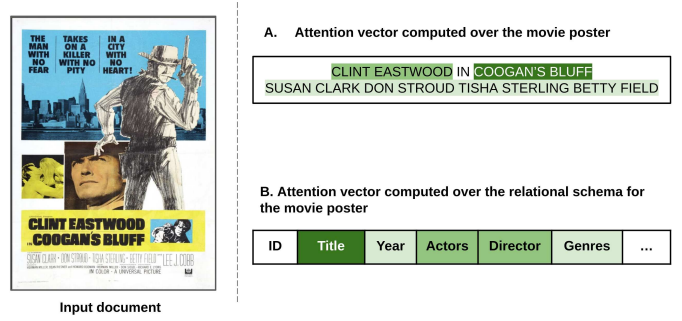}
    \caption{Visualization of bi-directional attention computed over a text-span in a movie poster and a relational tuple containing metadata about the movie. Darker shades refer to higher attention scores assigned by our network referring to higher likelihood of finding a match, whereas lighter shades refer to lower attention scores, signifying lower likelihood of finding a match. In this example, we observe higher probabilities of finding a matching tuple against the text spans ``Clint Eastwood'' and ``Coogan's Bluff'' against attributes \texttt{Actor, Director} and \texttt{Title} of a relational tuple in the database.\looseness=-1}\label{fig:2}
\end{figure}

\noindent \textbf{B.2. Pruning off unlikely matches: }Let's assume that the minimum distance between a text span and a cluster-centroid indexed in~$V_D$ is~$c$. We can compute this distance for all text spans appearing in the document to obtain a vector~$\Vec{C} = \{c_1,c_2...\}$. The probability of finding a matching tuple for a text span appearing in the document can therefore be defined as follows.\looseness=-1 

\begin{equation}
    \Vec{A_D} = 1 - \frac{\Vec{C}-min(\Vec{C})}{max(\Vec{C})-min(\Vec{C})}
\end{equation}

\noindent The~$i^{th}$ term in~$\Vec{A_D}$ represents the probability of finding a matching tuple for the~$i^{th}$ text span in the document. To discard those text spans that are unlikely to find a matching tuple in the database, we apply a filter on~$\Vec{A_D}$ using a mechanism called k-max weighted attention~\cite{bian2017compare}. It sorts the probability terms in~$\Vec{A}_D$ in descending order, retaining the top-k terms (k=25) and sets the rest to zero. The non-zero probability terms correspond to those text spans that are retained after the pruning stage is complete. Similarly, minimizing the distance between the embedding vector of the text span~$w$ projected onto the shared embedding space~(i.e. $\mathcal{F'}_w$) and the cluster-centroids indexed in~$V_T$ returns a vector~$\Vec{A}_T$ over the relational tuples. The~$i^{th}$ term in~$\Vec{A_T}$ represents the maximum probability of a text span being matched against the~$i^{th}$ attribute of a tuple in our database. To discard unlikely matches, we apply k-max weighted attention on~$\Vec{A}_T$, which retains top-k tuples (k=100) in~$\Vec{A}_T$ and sets the rest of the probability terms to zero. The non-zero terms in~$\Vec{A}_T$ correspond to those tuples that are retained after the pruning stage is complete. Fig.~\ref{fig:2} shows a visualization of the non-zero probability terms computed over a document and a relational tuple from one of our experimental datasets. Contrary to existing works that leverage carefully designed rules, we follow a principled, unsupervised learning technique to guide our pruning operation. Experiments show that the bi-directional attention scheme employed in our framework reduces the number of pairwise comparisons, which in turn reduces the end-to-end latency of our framework~(see Section~V.5)\looseness=-1\vspace{0.1cm}

\noindent \textbf{B.3. Aggregation: }We only consider those text spans (and relational tuples) that remain after the pruning step for pairwise comparisons (Eq.~1). Once we have identified a matching tuple for each text span in the input document, we can aggregate them from text span-level to document-level.\footnote{document-level aggregation is needed in many real-world applications where the number of matching tuples that can be returned for a document has a strict upper bound} The \textit{aggregation layer} executes this operation by performing majority voting amongst all matching tuples identified for all text spans in the document.\looseness=-1

\subsection{Training the Network}
\noindent We train our network using a learning objective similar to triplet loss~\cite{fernandez2019termite}. Our goal is to minimize the distance between a text span~$w$ \& its matching tuple~$t$ on their shared embedding space, and maximize the distance between~$w$ \& a non-matching tuple~$t'$ from the database. We formalize this learning objective as follows.   

\begin{multline}
    \mathbf{L}_{align}(w,t) = argmin_{\hspace{0.05cm}i,j} |\hspace{0.05cm} \mathcal{F'}_w[j] - \mathcal{F'}_t[i] \hspace{0.05cm}| - \\ \lambda\cdot argmin_{\hspace{0.05cm}i,j} |\hspace{0.05cm} \mathcal{F'}_w[j] - \mathcal{F'}_{t'}[i] \hspace{0.05cm}|
\end{multline}\label{eq:1}

\noindent The first term in Eq.~4 represents the minimum distance between a text span~$w$ and its matching tuple~$t$. The second term, on the other hand, represents the minimum distance between~$w$ and a non-matching tuple~$t'$, $\lambda$ is a hyperparameter. We set its value to 0.025. We obtain matching tuples for a text span from human-experts during training corpus construction. We train our network on a Tesla P100 GPU for 20 epochs using stochastic gradient descent with a batch size of 4. Each sample in our training corpus consists of a triplet, a text span, its matching tuple, and a non-matching tuple sampled from the database. We use early stopping to prevent overfitting and Adam optimizer with a learning-rate of $1\times 10^{-4}$, weight decay of $1\times 10^{-2}$ and $(\beta_1,\beta_2)$ = (0.9, 0.999) to train our network.\looseness=-1 

\subsection{End-to-end workflow}
\noindent Once our network is trained, we can identify matching tuples for a text span in two steps. In the first step, we filter out those tuples that are unlikely to match by computing their likelihood of finding a match against any text span in the document using the output from the attention layer of our network. We only keep the top-k most likely tuples from this step. The rest are discarded. In the second step, the remaining tuples are compared against the text span using Eq.~2, and the pairwise distance between each tuple and the text span is minimized. To find a matching tuple for an entire document, we follow the same steps as mentioned above, identify matching tuples for each word in the document, and then aggregate those results using the aggregation layer of our network.\looseness=-1 
\section{Experiments}
\noindent We seek to answer four key questions through our experiments: (a) \textit{how do we perform on entity matching tasks against heterogeneous documents?} (b) \textit{how do we compare against state-of-the-art baselines on each of these tasks?} (c) \textit{what are the individual contributions of some of the key components in our framework?}, and (d) \textit{is our framework computationally robust?}. We answer the first question by reporting the F1-score of two entity matching tasks on separate publicly available datasets (see Section~V.B.1). To answer the second question, we compare our performance against a number of state-of-the-art baselines in Section~V.B.4. We answer the third question through an ablation study in Section~V.B.5. Finally, we study the computational robustness of our framework by adapting it to a number of resource constrained environments and then reporting its performance on our experimental datasets in Section~V.B.6. We conduct all of our experiments on a system with 25GB RAM and a Tesla P100 GPU.\looseness=-1

\subsection{Experiment Design}

\noindent \textbf{A.1. Datasets:} We evaluate our framework on two real-world datasets. Each dataset contains documents with diverse layouts, formats, and textual content. We describe both of them below.\looseness=-1\vspace{0.05cm}

\begin{enumerate}[label=(\roman*)]

\item \textbf{IMDB Movie Dataset. }This dataset consists of approximately 8.4K movie posters from the IMDB website and a relational table collected from the IMDB movie database ({\url{https://datasets.imdbws.com/}}) containing an equal number of tuples. The table contains 12 unique attributes capturing various movie metadata, such as `Title', `Directors', `Actors' and more. The posters are stored as image files. The database is stored as a single JSON.\looseness=-1\vspace{0.1cm} 

\item \textbf{NYC Open Event Dataset. }This dataset consists of approximately 7.9K event publicity images curated by New York City Parks and Recreations Department. It also contains a relational table with 92.5K tuples from the NYC Open data website (\url{https://opendata.cityofnewyork.us/}). The table contains 24 unique attributes capturing various important event information, such as `Address', `Date', `Description' and more. The event posters are stored in image format and the database is stored as a JSON.\vspace{0.1cm}         
\end{enumerate}

\begin{figure}[b]
    \centering
    \includegraphics[width=\columnwidth]{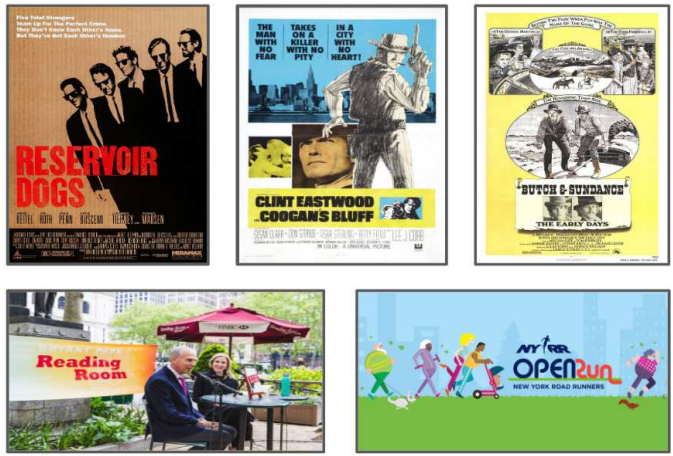}
    \caption{Sample documents from IMDB Movie Dataset (upper row) and NYC Open Event Dataset (bottom row)\looseness=-1}\label{fig:4_1}
\end{figure}

\noindent For both datasets, our objective is to map a document to its matching tuple in the database. We construct gold-standard labels for both tasks by annotating these matching pairs. Each pair consists of a text span in the document and its corresponding matching tuple in our database. We recruited 3 graduate students for this task and resolved all inter-annotator disagreements by consensus. We split both datasets into training, validation and test sets. Recall that a training sample in our experimental setup consists of a triplet -- a text span, its matching tuple, and a non-matching tuple randomly sampled from our database. The training and validation set consists of 2000 and 500 of such triplets respectively from each dataset. The rest of the samples comprise the test set. We have made both datasets available at: \url{https://github.com/anonsig2020/cem\_vrd}\looseness=-1\vspace{0.1cm}

\noindent \textbf{A.2. Evaluation metrics:} We compare a tuple inferred as a match by our framework for each document in our test corpus against its groundtruth label. We deem a match to be accurate if it has identified the same tuple as the groundtruth. We report the average precision and F1-score $@k=[1,5,20]$ for both datasets in Table~\ref{table:eval_summary}.\looseness=-1\vspace{0.1cm}

\noindent \textbf{A.3. Baselines:} We compare the downstream performance of our framework against a number of state-of-the-art baselines using the same experimental setup as described above. They are as follows.\looseness=-1\vspace{0.05cm}

\begin{enumerate}[label=(\roman*)]
    \item \textbf{Fuzzy string matching (M1). }In this unsupervised baseline method, we compare each attribute of a tuple in our database with n-grams in a visually rich document using fuzzy-string matching\footnote{\url{https://pypi.org/project/fuzzywuzzy/}}. This pairwise comparison results in a similarity score for every relational tuple in the database. We identify the matching tuple by selecting the tuple with the highest similarity score in the document.\looseness=-1\vspace{0.1cm}
    
    \item \textbf{Text embeddings (M2). }Following~\cite{nie_han_he_sun_chen_zhang_wu_kong_2019}, in this unsupervised baseline method we encode each word in the document as well as each attribute in a relational tuple using a pretrained RoBERTa\textsubscript{BASE} model. We identify a matching tuple in two steps. \textit{First}, we identify a matching tuple for each word using a k-nearest-neighbor-search over the entire database. \textit{Second}, we perform majority voting amongst all tuples retrieved from the first step to identify the tuple at the document-level.\looseness=-1\vspace{0.1cm}
    
    \item \textbf{Document IE (M3). }In this supervised baseline, we identify a matching tuple by performing a fuzzy-join operation. We extract a structured record from each document by employing the state-of-the-art LayoutLMV2\textsubscript{BASE} model~\cite{xu2020layoutlmv2} that is first pretrained on the IIT-CDIP dataset~\cite{harley2015evaluation} and then fine-tuned on our training corpus. To identify a matching tuple, we performing a fuzzy outer-join between the record and the relational database.\looseness=-1\vspace{0.1cm}
    
    \item \textbf{Graph-based embeddings (M4). }In this baseline method, we use \texttt{EMBDI}~\cite{cappuzzo2020creating} which is a state-of-the-art, graph-based approach to compute fixed-length representations of a relational tuple. We encode each word in the document using a pretrained RoBERTa\texttt{BASE} model. We identify the matching tuple for each document in two steps. \textit{First}, we identify a matching tuple for each word by solving an orthogonal Procrustes problem following Cappuzzo et al.~\cite{cappuzzo2020creating}. \textit{Second}, we perform majority voting amongst all tuples retrieved from the first step to identify the matching tuple at the document-level.\looseness=-1\vspace{0.1cm}
    
    \item \textbf{Hashing-based approach (M5). }Following Lin et al.~\cite{lin2015deep}, we construct a neural network that learns 8-bit binary hash-codes to represent each word in a document. We represent relational tuples as a sequence of tokens and compute 8-bit hash codes for them in a similar fashion. To identify the matching tuple for a word, we minimize the Hamming distance between two binary vectors, one representing the word itself and another one representing the tuple. We identify the matching tuple at document-level by performing majority voting amongst all tuples identified during word-level matching.\looseness=-1\vspace{0.1cm} 

    \item \textbf{Large language model (M6). }In this baseline, we employ the publicly available, pretrained GPT-Neo~\cite{gptneo} model to identify matching tuples from a relational database. We transcribe each document and feed the extracted text as an input to our model. For each tuple in our database, we prompt the model to output the following: if a tuple is semantically similar to the document, reply with a `yes' or `no'. If the answer to the previous question is `yes', assign a number between 1 to 10 to reflects the confidence in the accuracy of this answer. We use this setup to identify the matching tuple at document-level by sorting all the tuples identified from the previous step by their confidence scores and then returning the top-1 tuple.\looseness=-1\vspace{0.1cm}

    \item \textbf{Vision-language model (M7). }Following Radford et al.~\cite{radford2021learning}, in this baseline method we use a ResNet50-based image encoder model and a Transformer-based text encoder model simultaneously trained on \texttt{\{image, text\}} pairs collected via web crawling. We compute a fixed-length representation of each document from its rendered image using this image encoder model, and similarly for each relational tuple using the text encoder model. We identify the matching tuple for a document by minimizing the distance between these two vectors within our database. 
    
\end{enumerate}

\begin{table}[t]
  \caption{{End-to-end performance on experimental datasets}}\label{table:eval_summary}
  \renewcommand{\arraystretch}{1.1}
  \resizebox{\columnwidth}{!}
{\begin{tabular}{l|l|ccc}
    \toprule
    {\textbf{Dataset}} & {\textbf{k}} & {\textbf{Precision (\%)}} & {\textbf{F1 (\%)}}\\
    
    \midrule\midrule
    {IMDB Movie Dataset} & {k=1} & {{75.05}} & {{85.74}}\\
     & {k=5} & {{77.25}} & {{87.16}}\\
     & {k=20} & {{87.90}} & {{93.56}}\\
    \midrule
    {NYC Open Event Dataset} & {k=1} & {{58.80}} & {{74.05}}\\
     & {k=5} & {{60.10}} & {{75.07}}\\
     & {k=20} & {{68.06}} & {{80.99}}\\
  \bottomrule
\end{tabular}}
\end{table}

\subsection{Experimental Results}

\noindent \textbf{B.1. End-to-end performance: }We present the precision and F1-score $@k=[1,5,20]$ of our framework on both datasets in Table~\ref{table:eval_summary}. On the IMDB Movie Dataset, we obtain a top-1 precision of 75.05\% and a F1-score of 85.74\%, whereas for the NYC Open Event Dataset we obtain a top-1 precision of 56.80\% and a F1-score of 72.44\%. We obtain a perfect recall on both datasets as our framework returns a non-empty set of tuples for each document. Taking a closer look at these results reveals that we perform comparatively better on documents that are relatively verbose. The average turnaround latency of our framework is 0.8 secs/document for the IMDB Movie Dataset, and 3.1 secs/document for the NYC Open Event Dataset. We observe relatively higher turnaround latency for the NYC Open Event Dataset because the average number of candidate pairs participating in pairwise comparisons~(see Section~IV.B) is relatively higher for this dataset.\looseness=-1\vspace{0.05cm} 

\begin{figure}[t]
    \centering
    \includegraphics[width=\linewidth]{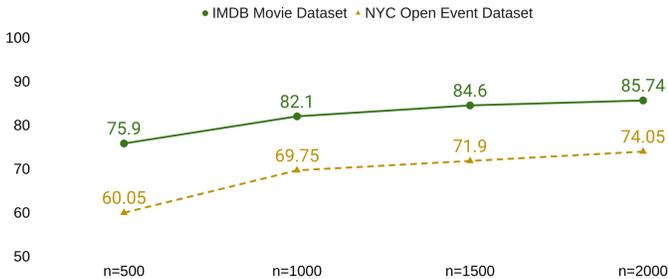}
    \caption{F1-score $@k$ = 1 for with $n=\{500,1000,1500,2000\}$ samples used to train our network\looseness=-1}\label{fig:5}
\end{figure}

\noindent \textbf{B.2. On the diversity of training samples: }We investigate the role played by the diversity of samples in our training corpus by varying the number of samples used to train our model. The rest of our experimental settings is kept unchanged. Results show~(see Fig.~\ref{fig:5}) that increasing the number of samples improves the average F1 score for both datasets. However, improvement in performance starts to plateau as the number of training samples exceeds a threshold.\looseness=-1\vspace{0.05cm}

\noindent \textbf{B.3. Label-efficiency: }We investigate the label-efficiency of our framework by training a model that has an identical architecture as ours but with one key difference. Recall that our network employs pretrained models (see Section IV.A) in its representation layer to leverage transfer learning~\cite{pan2009survey} and train in a label-efficient way. In this experiment, instead of these pretrained models, we use randomly initialized, trainable weights of same dimensions to encode each input sequence. We train this model using direct supervision with the same learning objective on the same training corpus. If its performance on the validation set is observed to be incomparable with our original model, we increase the size of its training corpus by introducing 25 additional samples. We keep increasing the number of samples in the training corpus in this way until we this baseline model has obtained a comparable F1 score with our original network.\looseness=-1 

\begin{table}[h]
  \caption{{Number of human-labeled samples needed for comparable downstream performance in a supervised setting}}\label{table:label_efficiency}
  \renewcommand{\arraystretch}{1.1}
  \resizebox{\columnwidth}{!}
{\begin{tabular}{l|cc}
    \toprule
    {\textbf{Dataset}} & {\textbf{\#No. of training samples}} & {\textbf{Saved (\%)}}\\
    
    \midrule\midrule
    {IMDB Movie Dataset} & 2000 & {{57.05}}\\
    \midrule
    {NYC Open Event Dataset} & 2000 & {{60.77}}\\
  \bottomrule
\end{tabular}}
\end{table}

Let, the number of human-labeled samples needed to train the baseline model for comparable performance is~$N_{\texttt{base}}$. The number of human-labeled samples needed to train our original network is~$N$. Therefore, we define the label-efficiency of our network as $\frac{N_{\texttt{base}} - N}{N}$. We report the number of human-labeled samples used to train our original network in the first column of Table~\ref{table:label_efficiency}. The second column in this table denotes the label-efficiency of our network against the supervised baseline on both datasets. Results show that our framework requires up to 60\% less human-labeled samples to obtain comparable performance than a fully supervised baseline.\looseness=-1\vspace{0.05cm}

\noindent \textbf{B.4. Comparison against baselines: }We compare our downstream performance against a number of baseline methods in Table~\ref{table:baselines}. The best performing models on both datasets are shown in boldface. 
We outperform the fuzzy-match based baseline (M1) by more than 37 F1 points on the IMDB Movie Dataset and 27 F1 points on the NYC Open Event Dataset. Diving deep at these results reveals that this string-matching based approach does not fare well on documents that have multiple potential matches in the database, and the matching tuple can only be disambiguated from contextual encodings. We observe similar trend against the text embedding-based baseline (M2) also. Results show that we outperform this baseline by more than 12 F1 points on the IMDB Movie Dataset and 8 F1 points on the NYC Open Event Dataset. This establishes the superiority of the multimodal encoding capability of our representation layer compared to off-the-shelf text-based embedding techniques.\looseness=-1 

\begin{table}[h]
  \caption{{End-to-end performance of all competing methods on our experimental datasets}}\label{table:baselines}
  \renewcommand{\arraystretch}{1.1}
  \resizebox{\columnwidth}{!}
{\begin{tabular}{l|l|ccc}
    \toprule
    {\textbf{Dataset}} & {\textbf{Method}} & {\textbf{Precision (\%)}} & {\textbf{F1 (\%)}}\\
    
    \midrule\midrule
    {IMDB Movie Dataset} & {Fuzzy matching (M1)} & {{33.33}} & {{48.72}}\\
     & {Text embeddings (M2)} & {{62.50}} & {{76.92}}\\
     & {Document IE (M3)} & {{72.33}} & {{82.32}}\\
     & {Graph-based (M4)} & {{66.50}} & {{79.87}}\\
     & {Hashing-based (M5)} & {{64.33}} & {{75.18}}\\
     & {Large Language Model (M6)} & {{73.18}} & {{82.47}}\\
     & {Vision-Language Model (M7)} & {{72.90}} & {{82.66}}\\
     & {Our method} & {\textbf{75.05}} & {\textbf{85.74}}\\
    \midrule
    {NYC Open Event Dataset} & {Fuzzy matching (M1)} & {{45.05}} & {{56.47}}\\
     & {Text embeddings (M2)} & {{45.95}} & {{62.54}}\\
     & {Document IE} (M3) & {{57.75}} & {{59.17}}\\
     & {Graph-based (M4)} & {{50.33}} & {{66.96}}\\
     & {Hashing-based (M5)} & {{48.60}} & {{63.25}}\\
     & {Large Language Model (M6)} & {{56.06}} & {{71.47}}\\
     & {Vision-Language Model (M7)} & {{55.80}} & {{72.25}}\\
     & {Our method} & {\textbf{58.80}} & {\textbf{74.05}}\\
  \bottomrule
\end{tabular}}
\end{table}

Compared to the Document IE-based baseline (M3), we observe an improvement of 2.72 F1 points for the IMDB Movie Dataset and 1.05 F1 points for the NYC Open Event Dataset. Training a document understanding model (e.g. LayoutLMv2) that extracts structured records from a visually rich document, however, requires additional human-labeling effort at the token-level to construct its training corpus. It is cumbersome and often hard to scale. We also outperform the graph-based baseline method (M4) on both datasets. Our results show that the semantic coherence of this graph-based encoding technique does not translate well for cross-modal entity matching tasks. 
We observe similar improvement in performance against the hashing-based baseline (M5) as well. Although the turnaround latency of this baseline is less than ours, we outperform it by more than 10 F1 points on both datasets. Finally, comparing the downstream performance of our framework against a pretrained GPT-Neo model (M6) reveals that we can outperform this baseline by 1.87 F1 points on the IMDB Movie Dataset and 2.74 F1 points on the NYC Open Event Dataset. Taking a closer look at these results reveal that this model tends to hallucinate on many documents in both datasets. It also requires additional steps to disambiguate among multiple potential matches for textually sparse documents. We also outperform the vision-language model-based baseline (M7) on both datasets. We hypothesize that this is because of relatively weaker representation capabilities of this baseline, stemming from its usage of pixel-level abstraction to represent each document using a vision encoder model. This establishes the importance of encoding both visual and textual features of a document for this entity matching task.\looseness=-1\vspace{0.05cm}

\begin{table}[t]
  \caption{{Results from the ablation study on IMDB Movie Dataset}}\label{table:ablation}
  \renewcommand{\arraystretch}{1.1}
  \resizebox{\columnwidth}{!}
{\begin{tabular}{l|c|c}
    \toprule
    {\textbf{Ablated component}} & {\textbf{Removed or Replaced?}} & {$\Delta$\textbf{F1 (\%)$\downarrow$}}\\
    
    \midrule\midrule
    {Attention layer} & {{Removed}} & {{2.24}}\\
    {Visual features} & {{Removed}} & {{4.90}}\\
    \midrule
    {Representation layer} & {{LayoutLMv2 replaced with RoBERTa}} & {{1.06}}\\
    {Representation layer} & {{MobileNet replaced with CLIP}} & {{-3.55}}\\
  \bottomrule
\end{tabular}}
\end{table}

\noindent \textbf{B.5. Ablation study: }To measure individual contributions of some of the key components in our framework, we perform an ablation study. In each of these ablative baselines, we remove or replace a key component in our framework and observe its effect on our downstream performance. We present our findings in Table~\ref{table:ablation}. The first column in this table specifies the component in our network that is being removed (or replaced), and the last column denotes the degradation in F1-score due to this change with respect to our original model. In the first ablative baseline, we remove the attention layer from our network. This results in a 2.24\% decrease in F1-score and $\approx 4$x increase in turnaround latency thus establishing the contribution of the bi-directional attention scheme employed by our network for fast alignment between a text span and a relational tuple. In the second baseline, we remove the visual features from the representation layer of our network. We observe that this results in a 4.90\% decrease in F1-score. This establishes the necessity of encoding both visual and textual features to represent a text span in a visually rich document. In our third ablative baseline, we replace the pretrained LayoutLMv2\textsubscript{BASE} model, which has been used to encode the textual features of a document in the representation layer of our network with a RoBERTa\textsubscript{BASE} model pretrained on the Google Book corpus. We observe a 1.06 F1 points drop in performance. This establishes the flexibility offered by our representation layer to plug-and-play stronger foundational models within our network in the future. In our final baseline method, we replace the pretrained MobileNet model with the same vision-language model~\cite{radford2021learning} employed by our final baseline method~(M7) to encode the visual features of the document. Results show that introducing an image encoder model that has been jointly trained on image and text pairs can benefit the representation capability of our model, improving its downstream performance. Investigating optimal ways to introduce even larger models within our network is one of our planned future works.\looseness=-1\vspace{0.05cm}

\noindent \textbf{B.6. Computational robustness: }One of the major challenges of deploying a deep entity matching framework for real-world applications is the amount of computational resources needed to infer a match. For instance, the cost of computing a fixed-length representation of a data element using a Transformer-based model with $d$ layers is: $O(n^2\cdot d + n\cdot d^2 + n\cdot d)$ for an input sequence length of~$n$. This quadratic cost makes it difficult to deploy these frameworks in resource-constrained environments without incurring any degradation in downstream performance. In \texttt{Juno}, the computational bottleneck stems from various pretrained models employed in the representation layer of our network (see Section~IV.A). Recent advances in controllable model compression~\cite{cai2019once, kim_length_adaptive_2021} makes it possible to mitigate this to some extent. They allow us to train a deep neural network only once, and then adapt it at run-time by setting certain model hyperparamters such as input sequence length, number of layers, based on computational constraints (e.g. working memory size, computational capability, number of cores etc.) imposed by the environment. We use publicly available algorithms officially released by their respective authors to adapt our network for various resource-constrained environments. We point interested readers to the original works by Cai et al.~\cite{cai2019once} and Kim et al.~\cite{kim_length_adaptive_2021} for more background on this.\looseness=-1 

\begin{table}[h]
  \caption{{End-to-end performance of our framework on various resource-constrained environments}}\label{table:robust}
  \renewcommand{\arraystretch}{1.1}
  \resizebox{\columnwidth}{!}
{\begin{tabular}{l|c|c|c}
    \toprule
    {\textbf{Environment}} & \textbf{$\Delta$ Model Footprint (\%)~$\downarrow$} & {\textbf{$\Delta$ GFlops (\%)~$\downarrow$}} & {\textbf{F1 (\%)}}\\
    
    \midrule\midrule
     A12 & 7.77 & 49.25 & 86.66\\
     A13 & {5.89} & 41.90 & {{86.95}}\\
     A14 & {24.36} & 68.75 & 87.0\\
     A15 & {20.74} & 56.33 & 86.95\\
     \midrule
     Original & -- & -- & 85.74\\
  \bottomrule
\end{tabular}}
\end{table}

\begin{table}[h]
  \caption{{Environmental constraints of various resource-constrained environments used in our experiments}}\label{table:environment}
  \renewcommand{\arraystretch}{1.1}
  \resizebox{\columnwidth}{!}
{\begin{tabular}{l|c|c|c|c}
    \toprule
    {\textbf{Environment}} & \textbf{Cores} & {\textbf{Threads}} & \textbf{Cache} & \textbf{Memory Capacity}\\
    
    \midrule\midrule
     A12 & 6 & 6 & 8MB & 4GB\\
     A13 & 6 & 6 & 8MB & 4GB\\
     A14 & 6 & 6 & 8MB & 6GB\\
     A15 & 6 & 6 & 12MB & 6GB\\
  \bottomrule
\end{tabular}}
\end{table}

In this section, we simulate such resource-constrained environments, adapt our framework to their computational constraints using~\cite{kim_length_adaptive_2021}, and report our downstream performance~(see Table~\ref{table:robust}). Each row in this table represents a computational environment that simulates a recently released iPhone. The left-most column in this table denotes the specific iPhone processor this computational environment simulates. We describe the computational constraints simulated in each of these environments in Table~\ref{table:environment}. We adapt our model based on these constraints using~\cite{kim_length_adaptive_2021} and report its performance on the IMDB Movie Dataset. The second and third column in Table~\ref{table:robust} represent the reduction in memory footprint and computational overhead of the resulting model compared to our original network. The final column in this table represents the downstream performance of the resulting model adapted for that environment. Our results show up to 24\% reduction in memory footprint and 68\% reduction in computational overhead of the original model without any degradation in downstream performance. This establishes the computational robustness of our framework across various resource-constrained environments, making it a suitable candidate for on-device deployment in edge-devices. Investigating optimal ways to adapt our framework for interactive applications in edge-devices is one of our planned future works.\looseness=-1

\section{Conclusion}
Visually rich documents are great sources of ad-hoc information. The information they contain, however, is often incomplete. This makes it difficult to contextualize the information retrieved from these documents and gather actionable insights from them. We develop \texttt{Juno} -- a generalizable framework to address this limitation by augmenting each document with supplementary information from a relational database. To identify matching tuples, we develop a multimodal neural network that maps text spans in the document to tuples in the database. Harnessing the power of pretrained models through transfer leaning, \texttt{Juno} executes this entity matching task with significantly less human-labeled samples. It ensures fast mapping against large-scale databases by leveraging a novel bi-directional attention mechanism that allows it to prune unlikely matches from the search space and reduce the number of pairwise comparisons. To the best of our knowledge, this is the first work that investigates the incompleteness of \texttt{VRDs} and proposes a \textit{generalizable, performant and computationally robust} framework to address it in an end-to-end way. Contrary to existing works, \texttt{Juno} does not utilize any handcrafted rules, or prior knowledge about the document type and/or the underlying schema to achieve its goal. Experiments on two heterogeneous datasets for separate entity matching tasks show that it is not only \textit{more performant} than state-of-the-art baseline methods -- outperforming them by more than 6\% in F1-score, but also more scalable -- reducing the number of human-labeled samples needed to train a supervised baseline with the same backbone that achieves comparable performance by up to 60\%. Our experiments also show that \texttt{Juno} is computationally robust. Contrary to existing vision-language models, we can use off-the-shelf algorithms to adapt it for resource-constrained environments, reducing its memory footprint by up to 24\% without any performance degradation on real-world datasets. Investigating an optimal way to adapt deep entity matching frameworks for interactive applications on edge-devices~\cite{burley2022quill} is one of our planned future works.\looseness=-1

\balance

\bibliographystyle{IEEEtran}
\bibliography{references}

\end{document}